\documentclass{article}[10pt]

\usepackage{url}
\usepackage[dvips]{graphicx}

\begin{document}
\title{\Large The Generalized Traveling Salesman Problem \\solved with Ant Algorithms}

\author{\small{Camelia-M. Pintea, Petric\u a C. Pop, Camelia Chira}\\
  \small {North University Baia Mare, Babes-Bolyai University, Romania}\\
{\small cmpintea@yahoo.com, pop\_petrica@yahoo.com, cchira@cs.ubbcluj.ro}\\
}
\date{\vspace{-5ex}}
 \maketitle
\small
\sffamily

\begin{abstract}
A well known ${\cal NP}$-hard problem called the {\em Generalized Traveling Salesman Problem (GTSP)} is considered. In {\em GTSP} the nodes of a complete undirected graph are partitioned
into clusters. The objective is to find  a minimum cost tour passing through exactly one node from each cluster. 

An exact exponential time algorithm and an effective meta-heuristic algorithm for the problem are presented. The meta-heuristic proposed is a modified {\em Ant Colony System} ({\em ACS}) algorithm called {\em Reinforcing Ant Colony System} ({\em RACS}) which introduces new correction rules in the {\em ACS} algorithm. 
Computational results are reported for many standard test problems. The proposed  algorithm is competitive with the other already proposed heuristics for the {\em GTSP} in both solution quality and computational time.
\end{abstract}

\section{Introduction}

Many combinatorial optimization problems are ${\cal NP}$-hard, and the theory of ${\cal NP}$-completeness has reduced hopes that ${\cal NP}$-{\it hard} problems can be solved within polynomial bounded computation times. Nevertheless, sub-optimal solutions are sometimes easy to find. Consequently, there is much
interest in approximation and heuristic algorithms that can find near optimal solutions within reasonable running time. Heuristic algorithms are typically among the best strategies in terms of efficiency and solution quality for problems of realistic size and complexity.

In contrast to individual heuristic algorithms that are designed to solve a specific problem, meta-heuristics are strategic problem solving frameworks that can be adapted to solve a wide variety of problems. Meta-heuristic algorithms are widely recognized as one of the most practical approaches for combinatorial optimization problems. The most  representative meta-heuristics include genetic algorithms, simulated annealing, tabu search and ant colony. Useful references regarding meta-heuristic methods can be found in \cite{gk}.

{\em The Generalized Traveling Salesman Problem (GTSP)} has been introduced in \cite{ln} and \cite{nb}. The {\em GTSP} has several applications to location and 
telecommunication problems. More information on these problems and their applications can be found in \cite{figo,fist,ln}.

Several approaches were considered for solving the {\em GTSP}: a branch-and-cut algorithm for {\em Symmetric GTSP} is described and analyzed in \cite{fist}, in \cite{nb} is given a Lagrangian-based approach for {\em Asymmetric GTSP}, in \cite{snda} is described a random-key genetic
algorithm for the {\em GTSP}, in \cite{rb} it is proposed an efficient composite heuristic for the {\em Symmetric GTSP} etc.

The aim of this paper is to provide an exact algorithm for the {\em GTSP} as well as an effective meta-heuristic algorithm for the problem. The proposed  meta-heuristic is a modified version of {\em Ant Colony System (ACS)}. Introduced in (\cite{cdm,do}), {\em Ant System} is a heuristic algorithm inspired by the observation of real ant colonies. {\em ACS} is used to solve hard combinatorial optimization problems including the {\em Traveling Salesman Problem (TSP)}.

\section{Definition and complexity of the GTSP}

Let $G=(V,E)$ be an $n$-node undirected graph whose edges are associated with non-negative costs. We will assume w.l.o.g. that $G$ is a complete graph (if there is no edge between two nodes, we can add it with an infinite cost). 

Let $V_1,...,V_p$ be a partition of $V$ into $p$ subsets called {\it clusters} (i.e.
$V=V_1 \cup V_2 \cup ... \cup V_p$ and $V_l \cap V_k = \emptyset$
for all $l,k \in \{1,...,p\}$). We denote the cost of an edge
$e=\{i,j\}\in E$ by $c_{ij}$.

The {\it generalized traveling salesman problem} ({\em GTSP}) asks for finding a minimum-cost tour $H$ spanning a subset of nodes such that $H$ contains exactly one node from each cluster $V_i$, $i\in \{1,...,p\}$. 
The problem involves two related decisions:  choosing a node subset $S\subseteq V$, such that $|S \cap
V_k | = 1$, for all $k=1,...,p$ and  finding a minimum cost Hamiltonian cycle in the subgraph of
$G$ induced by $S$.

Such a cycle is called a  {\it Hamiltonian tour}. The {\em GTSP} is called {\em symmetric} if and only if the equality $c(i,j)=c(j,i)$ holds for every $i,j \in V$, where $c$ is the cost function associated to the edges of $G$.

\section{An exact algorithm for the GTSP}

In this section, we present an algorithm that finds an exact solution to the {\em GTSP}.

Given a sequence $(V_{k_{1}},...,V_{k_{p}})$ in which the clusters
are visited, we want to find the best feasible Hamiltonian tour
$H^*$ (w.r.t cost minimization), visiting the clusters according to the given sequence. This can be done in polynomial time by solving $|V_{k_{1}}|$ shortest path problems as described below. 

We construct a layered network, denoted by LN, having $p+1$ layers
corresponding to the clusters $V_{k_{1}},...,V_{k_{p}}$ and in
addition we duplicate the cluster $V_{k_{1}}$. The layered network
contains all the nodes of $G$ plus some extra nodes $v'$ for each
$v\in V_{k_1}$. There is an arc $(i,j)$ for each $i\in V_{k_l}$
and $j\in V_{k_{l+1}}$ ($l=1,...,p-1$), having the cost $c_{ij}$
and an arc $(i,h)$, $i,h \in V_{k_l}$, ($l=2,...,p$) having cost
$c_{ih}$. Moreover, there is an arc $(i,j')$ for each $i\in
V_{k_p}$ and $j'\in V_{k_1}$ having cost $c_{ij'}$.

For any given $v\in V_{k_1}$, we consider paths from $v$ to $w'$,
$w'\in V_{k_1}$, that visits exactly two nodes from each cluster
$V_{k_{2}},...,V_{k_{p}}$, hence it gives a feasible Hamiltonian
tour.

Conversely, every Hamiltonian tour visiting the clusters according
to the sequence $(V_{k_{1}},...,V_{k_{p}})$ corresponds to a path
in the layered network from a certain node $v\in V_{k_1}$ to
$w'\in V_{k_1}$.

Therefore the best (w.r.t cost minimization)
Hamiltonian tour $H^*$ visiting the clusters in a given sequence
can be found by determining all the shortest paths from each $v\in
V_{k_1}$ to each $w'\in V_{k_1}$ with the property that visits
exactly one node from cluster.

The overall time complexity is then $|V_{k_1}|O(m+n\log n)$, i.e.
$O(nm+nlogn)$ in the worst case. We can reduce the time by
choosing $|V_{k_1}|$ as the cluster with minimum cardinality.

It should be noted  that the above procedure leads to an $O((p-1)!(nm+nlogn))$ time exact algorithm for the {\em GTSP}, obtained by trying all the $(p-1)!$ possible cluster sequences. Therefore we have established the following result: the above procedure provides an exact solution to the {\em GSTP} in $O((p-1)!(nm+nlogn))$ time, where $n$ is the number of nodes, $m$ is the number of edges and $p$ is the number of clusters in the input graph. Clearly, the algorithm presented is an exponential time algorithm unless the number of clusters $p$ is fixed.

\section{Ant Colony System}

{\em Ant System} proposed in  \cite{cdm,do} is a multi-agent approach used for various combinatorial optimization problems. The algorithms were inspired by the observation of real ant colonies. 

An ant can find shortest paths between food sources and a nest. While walking from food sources to the nest and vice versa, ants deposit on the ground a substance called pheromone, forming a pheromone trail. Ants can smell pheromone and, when choosing their way, they tend to choose paths marked by stronger pheromone concentrations. It has been shown that this pheromone trail following behavior employed by a colony of ants can lead to the emergence of shortest paths. 

When an obstacle breaks the path ants try to get around the obstacle randomly choosing either way. If the two paths encircling the obstacle have the different length, more ants pass the shorter route on their continuous pendulum motion between the nest points in particular time interval. While each ant keeps marking its way by pheromone the shorter route attracts more pheromone concentrations and consequently more
and more ants choose this route. This feedback finally leads to a stage where the entire ant colony uses the shortest path. There are many variations of the ant colony optimization applied on various classical problems. {\em Ant System} make use of simple agents called ants which iterative construct candidate solution to a combinatorial optimization problem. The ants solution construction is guided by
pheromone trails and problem dependent heuristic information. 

An individual ant constructs candidate solutions by starting with
an empty solution and then iterative adding solution components until a complete candidate solution is generated. Each point at
which an ant has to decide which solution component to add to its
current partial solution is called a choice point. After the solution construction is completed, the ants give
feedback on the solutions they have constructed by depositing
pheromone on solution components which they have used in their
solution. Solution components which are part of better solutions
or are used by many ants will receive a higher  amount of
pheromone and, hence, will more likely be used by the ants in
future iterations of the algorithm. To avoid the search getting
stuck, typically before the pheromone trails get reinforced, all pheromone trails are decreased by a factor.

\textit{Ant Colony System} ({\em ACS}) was developed to improve
\textit{Ant System}, making it more efficient and robust. \textit{Ant Colony System} works as follows:
 \textit{m} ants are initially positioned on n  nodes chosen according to some initialization rule, for example randomly. Each ant builds a tour  by repeatedly applying a stochastic greedy rule - the state transition rule.
While constructing its tour, an ant also modifies the amount
of  pheromone   on the visited edges by applying the local
updating rule. Once all ants have terminated their tour, the
amount of pheromone on edges is modified again by applying the
global updating rule.  As was the case in ant system, ants are
guided, in building their tours by both heuristic information and by pheromone information: an edge with a high amount of pheromone is a very desirable
choice. The pheromone updating rules are designed so that
they tend to give more pheromone to edges which should be visited
by ants.

The ants solutions are not guaranteed to be optimal with respect to local changes and hence may be further improved using local search methods. Based on this observation, the best performance are obtained using hybrid algorithms combining probabilistic solution construction by a colony of ants with local search algorithms as 2-3 opt, tabu-search etc. In such hybrid algorithms, the ants can be seen as guiding the local search by constructing promising initial solutions, because ants preferably use solution components which, earlier in the search, have been contained in good locally optimal solutions.

\section{Reinforcing Ant Colony System for GTSP}

An {\em Ant Colony System} for the {\em GTSP} it is introduced. In order to enforces the construction of a valid solution used in {\em ACS} a new algorithm called {\em Reinforcing Ant Colony System} {\em (RACS)} it is elaborated   with a new pheromone rule as in \cite{cdd} and pheromone evaporation technique as in \cite{th}.

Let  $V_k(y)$ denote the node $y$ from the cluster $V_k$. The {\em RACS}
algorithm for the {\em GTSP} works as follows:

\begin{itemize}
\item Initially the ants are placed in the  nodes of the graph, choosing randomly the \textit{clusters} and also a random node from the chosen cluster

\item At iteration $t+1$ every ant moves  to a new node from an unvisited \textit{cluster} and the parameters controlling the algorithm are updated.

\item  Each edge is labeled by a trail intensity. Let
$\tau_{ij}(t)$ represent the trail intensity of the edge $(i,j)$ at time $t$. An ant decides which node is the next move with a probability that is based on the distance to that node (i.e. cost of the edge) and the amount of trail intensity on the connecting edge. The inverse of distance from a node to the next
node is known as the \textit{visibility},
$\eta_{ij}=\frac{1}{c_{ij}}$.

\item At each time unit evaporation takes place. This is to stop the intensity trails increasing unbounded. The rate evaporation is denoted by $\rho$, and its value is between 0 and 1. In order to
stop ants visiting the same \textit{cluster} in the same tour a tabu list is maintained. This prevents ants visiting \textit{clusters} they have previously visited. The ant tabu list is cleared after each completed tour.

\item To favor the selection of an edge that has a high pheromone value, $\tau$, and high visibility value, $\eta$ a probability function ${p^{k}}_{iu}$ is considered. ${J^{k}}_{i}$ are the unvisited neighbors of node $i$ by ant $k$ and $u\in {J^{k}}_{i},
u=V_k(y)$, being the node $y$ from the unvisited cluster $V_k$. This probability function is defined as follows:

\begin{equation}\label{probabilitate}
{p^{k}}_{iu}(t)= \frac{[\tau_{iu}(t)] [\eta_{iu}(t)]^{\beta}}
{\Sigma_{o\in {J^{k}}_{i}}[\tau_{io}(t)] [\eta_{io}(t)]^{\beta}} ,
\end{equation}

\indent where $\beta$ is a parameter used for tuning the relative
importance of edge cost in selecting the next node. ${p^{k}}_{iu}$ is the probability of choosing $j=u$, where  $u=V_k(y) $ is the next node, if $q>q_{0}$ (the current node is $i$). If $q\leq q_{0}$ the next node $j$ is chosen as follows:
\begin{equation}
j=argmax_{u\in J^{k}_{i}} \{\tau_{iu}(t)
{[\eta_{iu}(t)]}^{\beta}\} ,
\end{equation}
\noindent where $q$ is a random variable uniformly distributed over $[0,1]$ and $q_{0}$ is a parameter similar to the temperature in simulated annealing, $0\leq q_{0}\leq 1$.

\item After each transition the trail intensity is updated using the correction rule from \cite{cdd}:

\begin{equation}
\tau_{ij}(t+1)=(1-\rho)\tau_{ij}(t)+\rho \frac{1}{n \cdot L^{+}} .
\end{equation}

where $L^{+}$ is the cost of the best tour.

\item In {\em Ant Colony System} only the ant that generate the best tour is allowed to \textit{globally} update the pheromone. The global update rule is applied to the edges belonging to the {\it best tour}. The correction rule is

\begin{equation}\label{global}
\tau_{ij}(t+1)=(1-\rho) \tau_{ij}(t)+\rho  \Delta \tau_{ij}(t) ,
\end{equation}

\noindent where $\Delta\tau_{ij}(t)$ is the inverse cost of the best tour.

\item In order to avoid stagnation we used the pheromone
evaporation technique introduced in \cite{th}.
When the pheromone trail is over an upper bound $\tau_{max}$, the pheromone trail is re-initialized. The pheromone evaporation is used after the global pheromone update rule.

\end{itemize}

The {\em RACS} algorithm computes for a given time $time_{max}$ a sub-optimal solution, the optimal solution if it is possible and can be stated as follows in the pseudo-code description.
\begin{figure}
\centering \includegraphics[scale=0.9]{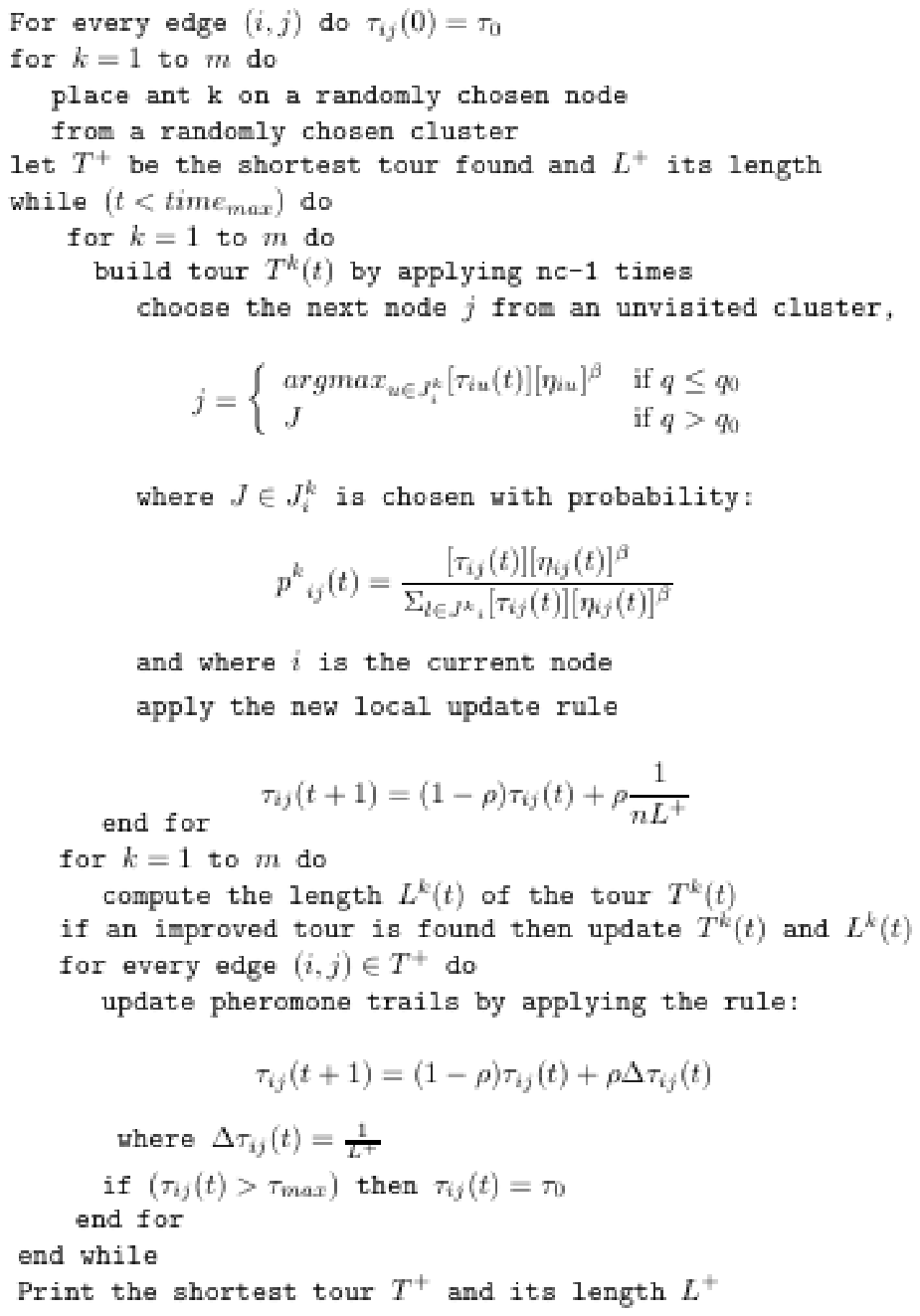}
\caption{\small \sffamily Pseudo-code: the {\em Reinforcing Ant Colony System} ({\em RACS}).}
\label{fig:figura11}
\end{figure}

\begin{figure}
\centering \includegraphics[scale=0.55]{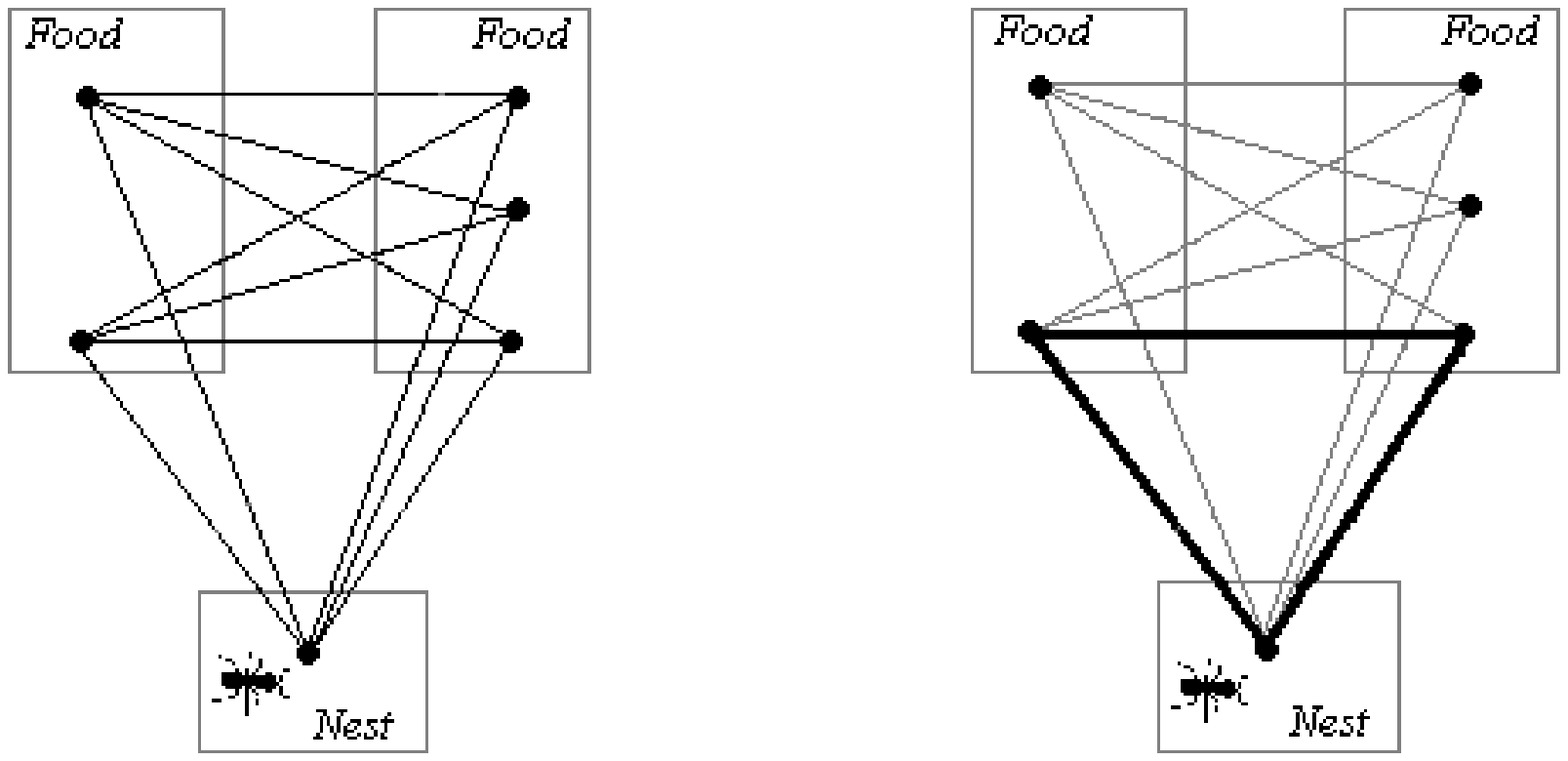}
\caption{\small \sffamily A graphic representation of the {\em Generalized Traveling Salesman Problem} ({\em GTSP})  solved with an ant-based heuristic called {\em Reinforcing Ant Colony System} ({\em RACS}) is illustrated. The first picture shows an ant starting from the nest to find food, going once through each cluster and returning to the nest; all the ways are initialized with the same $\tau_{0}$ pheromone quantity; after several iterations performed by each ant from the nest, the solution is 
visible.  The second picture shows a solution of {\em Generalized Traveling Salesman Problem} ({\em GTSP}) represented by the  largest pheromone trail (thick lines); the pheromone is evaporating on all the other trails (gray lines). 
}
\label{fig:figura}
\end{figure}

\section{Representation and computational results}

A graphic representation of {\em Reinforcing Ant Colony System} for solving {\em GTSP} is show in Fig.~\ref{fig:figura}. At the beginning, the ants are in their nest and will start to search food in a specific area. Assuming that each cluster has specific food and the ants are capable to recognize this, they will choose each time a different cluster. The pheromone trails will guide the ants to the shorter path, a solution of {\em GTSP}, as in Fig.~\ref{fig:figura}.

To evaluate the performance of the proposed algorithm, the {\em RACS} was compared to the basic {\em ACS} algorithm for {\em GTSP} and furthermore to other heuristics from literature: {\em Nearest Neighbor (NN)}, a composite heuristic $GI^{3}$ and a  {\em random key-Genetic Algorithm} \cite{rb,snda}. 
The numerical experiments that compare {\em RACS} with other heuristics used problems from {\em TSP} library \cite{tl}. {\em TSPLIB} provides optimal objective values for each of the problems. Several problems with Euclidean distances have been considered. The exact algorithm proposed in section 3, is clearly outperformed by heuristics including {\em RACS}, because his running time is exponential, while heuristics including {\em RACS} are polynomial time algorithms and provide good sub-optimal solution for reasonable sizes of the problem. 

To divide the set of nodes into subsets we used the procedure proposed in \cite{figo}.
This procedure sets the number of clusters $m=[n/5]$, identifies
the $m$ farthest nodes from each other, called centers, and
assigns each remaining node to its nearest center. Obviously, some
real world problems may have different cluster structures, but the
solution procedure presented in this paper is able to handle any
cluster structure. 

The initial value of all pheromone trails, $\tau_{0}=1/(n \cdot L_{nn})$, where $L_{nn}$ is the result of \textit{Nearest Neighbor, (NN)} algorithm. In {\em NN} algorithm the rule is always to go next
to the nearest as-yet-unvisited location. The corresponding tour traverses the nodes in the constructed order. For the pheromone evaporation phase, let denote the upper bound with $\tau_{max}=1/((1-\rho) L_{nn})$.

The decimal values can be treated as parameters and can be  changed if it is necessary. The parameters for the algorithm are critical as in all other ant systems. 

Currently there is no mathematical analysis developed to give the optimal parameter in each situation. In the {\em ACS} and {\em RACS} algorithm the values of the parameters were chosen as follows: $\beta=5$, $\rho=0.5$, $q_{0}=0.5$. 

\begin{figure}
\centering \includegraphics[scale=0.6]{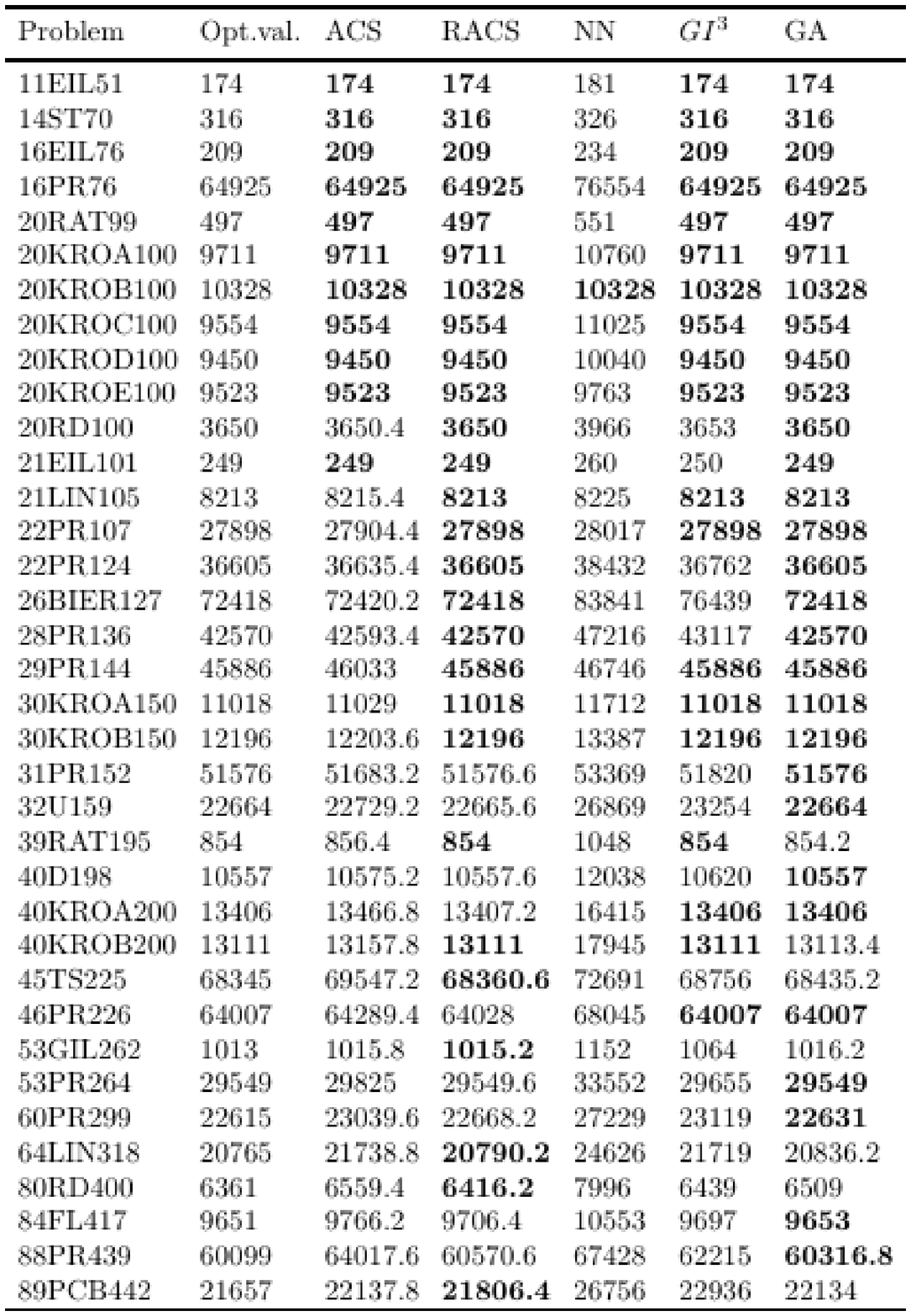}
\caption{\small \sffamily Reinforcing Ant Colony System (RACS) versus ACS, NN, G$I^{3}$ and GA.}
\label{fig:figura12}
\end{figure}

In  table from Figure~\ref{fig:figura12} we compare the computational results for solving
the {\em GTSP} using the {\em ACS} and {\em RACS} algorithm with the computational results
obtained using {\em NN}, $GI^{3}$ and {\em random key-Genetic Algorithm} mentioned above.  

The columns in table from Figure~\ref{fig:figura12} are as follows: {\bf Problem}- the name of the test problem; the first digits give the number of clusters ($nc$)and the last ones give the number of nodes ($n$); {\bf Opt.val.}-the optimal objective value for the problem \cite{snda}; {\bf ACS, RACS, NN, G$I^{3}$, GA}- the objective value returned by {\em ACS, RACS, NN, $GI^{3}$} and {\em random-key Genetic Algorithm}.

The best solutions are in Table ~\ref{fig:figura12} in the bold format. All the solutions of {\em ACS} and {\em RACS} are the average of five successively run of the algorithm, for each problem.  Termination criteria for {\em ACS} and {\em RACS} is  given by the $time_{max}$ the maximal computing time set by the user; in this case  ten  minutes. 

Table ~\ref{fig:figura12} shows  that {\em Reinforcing Ant Colony System} performed well finding the optimal solution in many cases. The results of {\em RACS} are better than the results of {\em ACS}. The {\em RACS} algorithm for the {\em Generalized Traveling Salesman Problem} can be improved if more appropriate values for the parameters are used. Also, an efficient combination with other algorithms can potentially improve the results.
\newpage 
\section{Conclusion}
The basic idea of {\em ACS} is that of simulating the behavior of a set of agents that cooperate to solve an optimization problem by means of simple communications.  The algorithm introduced to solve the {\em Generalized Traveling Salesman Problem}, called {\em Reinforcing Ant Colony System}, is an {\em ACS}-based algorithm with new correction rules. 

The computational results of the proposed {\em RACS} algorithm are good and competitive in both solution quality and computational time with the existing heuristics from the literature \cite{rb,snda}. The {\em RACS} results can be improved by considering better values for the parameters or combining the {\em RACS} algorithm with other optimization algorithms. Some disadvantages have also been identified and they refer the multiple parameters used for the algorithm and the high hardware resources requirements.

\end{document}